# SCENE: Evaluating Explainable AI Techniques Using Soft Counterfactuals

Haoran Zheng[1]

Physical Sciences Division,

The University of Chicago

Utku Pamuksuz (Faculty Advisors)

Physical Sciences Division,

The University of Chicago

**Abstract**

Explainable Artificial Intelligence (XAI) plays a crucial role in enhancing the transparency and accountability of AI models, particularly in natural language processing (NLP) tasks. However, popular XAI methods such as *LIME* and *SHAP* have been found to be unstable and potentially misleading, underscoring the need for a standardized evaluation approach. This paper introduces SCENE *(Soft Counterfactual Evaluation for Natural language Explainability)*, a novel evaluation method that leverages large language models (LLMs) to generate *Soft Counterfactual* explanations in a zero-shot manner.[2] By focusing on token-based substitutions, SCENE creates contextually appropriate and semantically meaningful Soft Counterfactuals without extensive fine-tuning. SCENE adopts *Validity$_{soft}$* and *C$_{soft}$* metrics to assess the effectiveness of model-agnostic XAI methods in text classification tasks. Applied to CNN, RNN, and Transformer architectures, SCENE provides valuable insights into the strengths and limitations of various XAI techniques.

## 1 Introduction

The majority of literature describes counterfactual explanations as "the smallest change to the feature values that changes the prediction to a predefined output" (Molnar, 2020, p. 263), or, in other words, what could have happened differently. Its biggest advantage is being very "human-friendly," as it requires neither additional assumptions nor complex operations in the background. (Molnar, 2020)

In this approach, specific values in an input instance are perturbed while all other variables remain unchanged. This process aims to see how these adjustments impact the outcome. Any perturbation that leads to an output change is identified as a counterfactual. By comparing the original instance with these counterfactuals, we gain insights into the relationship between inputs and outputs. (Wachter et al., 2017; Robeer et al., 2021). Furthermore, in the case of text data, particularly in classification tasks, perturbations usually involve replacing significant words or tokens with other highly significant ones that correspond to a different output class (Robeer et al., 2021; Wu et al., 2021).

The fundamental challenge for the field of causal inference, as highlighted by Holland (1986), is that it is impossible to observe both original instance and its alternative simultaneously for a single unit of analysis. This unit of analysis refers to the smallest entity about which we aim to make counterfactual investigations. This challenge renders causal inference more complex than statistical inference and unachievable without making certain identification assumptions (Feder et al., 2022).

Although counterfactual predictions are typically unobservable, in scenarios where the causal system is the predictor (the text sequence classifier in our case) itself, it is feasible to generate counterfactuals. In the context of this paper, which focuses on text classification tasks that fit these criteria, we can measure treatment effects by comparing predictions under both factual and counterfactual conditions (Feder et al., 2022). Consequently, this approach enables us to evaluate the effectiveness of existing explainable AI (XAI) techniques in identifying significant tokens.

---

[1] haoranzheng@uchicago.edu

[2] All experimentation and code can be found at https://github.com/HaoranZhengRaul/SCENE.git

In this paper, we introduce *SCENE (Soft Counterfactual Evaluation for Natural language Explainability)*, a novel approach that evaluates model-agnostic XAI techniques in natural language processing (NLP) tasks by leveraging large language models (LLMs) in a zero-shot manner to generate counterfactuals.

## 2 Dataset Description and Task Overview

The data selected for this analysis is the Stanford Sentiment Treebank (SST2) (Zaidan et al., 2007). This dataset consists of realistic movie reviews with binary labels, positive and negative. The complexity of the dataset, with an average review length of 773 words, mirrors real-world textual data. With its detailed labels and diverse linguistic styles, SST2 is ideal for benchmarking sentiment analysis algorithms. In this study, we focus on sentiment analysis as a text classification task, leveraging the annotations and varied linguistic expressions within SST2 to evaluate the performance of various model-agnostic XAI techniques.

## 3 Evaluation Metrics for XAI Techniques

### 3.1 Previous Works

The most prevalent method for explaining the outputs of NLP models involves assigning importance scores to individual input tokens. This approach provides a straightforward means of quantifying saliency explanations, summarizing model behavior through perturbation-based XAI approaches, attention-based XAI methods, or back-propagation-based XAI techniques (Wu et al., 2021). However, token importance scores are not without limitations. Perturbation-based XAI methods such as *LIME* and *SHAP*, while popular, are particularly vulnerable to adversarial attacks, leading to potentially misleading token importance scores (Slack et al., 2020). Furthermore, LIME and SHAP approximate importance by masking tokens, which may fail to accurately capture the model's behavior in natural counterfactual scenarios. This can result in situations where substituting seemingly insignificant tokens alters model predictions (Wu et al., 2021). Consequently, there is a pressing need for a rigorous and standardized framework to evaluate the performance of XAI techniques.

Significant contributions have been made in previous works towards developing metrics for evaluating saliency explanations. One common approach involves erasure-based methods, where significant tokens are masked or pruned, and the original model outputs are compared with the post-erasure outputs holistically. Notable examples include the work by Serrano and Smith (2019), which measured the percentage of decision changes after zeroing out the attention weights of "important" elements. Building on this concept, DeYoung et al. (2019) introduced the metrics of *Comprehensiveness* and *Sufficiency* to systematically evaluate erasure-based methods. These metrics quantify the impact on the model's predictions when the most important features are excluded or solely included.

However, previous works have raised some concerns with erasure-based approach. Feng et al. (2018) reveal not only that neural models can still make high-confidence predictions on reduced inputs, but also that these inputs are nonsensical to human observers, thereby highlighting interpretability issues. Intuitively, consider classifying the sentence "I love this movie" in a sentiment analysis task. If the token "love" is stripped or masked, resulting in "I [MASK] this movie," the sentence becomes incomplete and lacks actual sentiment and semantic meaning. These "damaged" versions of the inputs can fall outside the data distribution of the model's training sets, leading to imprecise evaluations of faithfulness (Ge et al., 2021).

Another fundamental sanity check is *the completeness axiom*, also known as the *Summation-to-Delta Property*, as established by Shrikumar et al. (2017) and Sundararajan et al. (2017). This axiom asserts that the sum of the attributions (explanations) equals the difference between the model's prediction for the original input and a baseline. This principle ensures that the explanation fully accounts for the model's decision, providing a comprehensive and theoretically sound foundation for attributing importance scores to individual tokens. Intuitively, this means that if the model's prediction changes significantly when a token is perturbed, the token's attribution should accurately reflect this change, adhering to the principle of completeness.

Inspired by the completeness axiom, the metric of *infidelity* was introduced (Yeh et al., 2019). Infidelity measures the discrepancy between the predicted impact of perturbations on the model's output and the actual impact observed. In the context of sentiment analysis, which is the focus of this paper, the goal is to identify the most important tokens influencing sentiment.

$$E_{I \sim \mu_I} \left[ \left( I^T \, \Phi(f,x) - \big(f(x) - f(x-I)\big) \right)^2 \right]$$

Where:

$f$ is a black box function
$\Phi$ is the XAI technique
$I$ is a random variable with probability measure $\mu_I$

To recreate the experiment, we implicitly use the original input text embeddings as the baseline. The perturbation function is designed to introduce small random noise to these embeddings, thereby creating perturbed versions of the input. Specifically, noise sampled from a normal distribution is subtracted from the original embeddings.

While this mathematical operation on word embeddings is theoretically sound and highly effective for computational purposes, it results in high-dimensional vectors composed of embeddings that often do not correspond to real words. Consequently, this poses challenges for human observers in semantically interpreting how specific perturbations affect the model's predictions. Furthermore, although to a lesser degree, the concern of plausibility remains, as the perturbed versions of the inputs can still fall outside the original data distribution.

Despite these challenges, using the original input embeddings as the baseline has the advantage of maintaining the context and structure of the input text, ensuring that the perturbations are applied in a meaningful manner. Since the evaluation of infidelity involves comparing the changes in the model's output due to these perturbations with the attributions provided by the saliency explanation, a lower infidelity score indicates that the explanation method accurately captures the importance of the tokens in the input text, thus validating the effectiveness of the saliency explanations.

Lastly, human agreement is often used as a benchmark to evaluate saliency explanations (DeYoung et al., 2019; Atanasova, 2024). Human agreement is measured by how well the identified tokens align with those annotated by humans. For models that make binary (yes/no) decisions about the relevance of saliency explanations, simpler metrics like accuracy or exact match are sufficient. In our case, most of the XAI methodologies generate continuous attribute importance scores, for which metrics such as the area under the precision-recall curve (AUPRC) and average precision, which evaluate the overlap between the model's extracted important tokens and the human-provided ones, are more appropriate. In this paper, we chose to use the human annotations collected by DeYoung et al. (2019) and the mean of the average precisions (MAP) to measure the performance of various XAI techniques (Atanasova, 2024).

$$MAP = \frac{1}{N} \sum_n \sum_\alpha (R_{n,\alpha} - R_{n,\alpha-1}) P_{n,\alpha}$$

Where:
- $N$ is the number of instances
- $P_{n,\alpha}$ and $R_{n,\alpha}$ are the precision and recall of the $\alpha$-th threshold and the $n$-th instance

## 3.2 Counterfactuals as an Alternative

### 3.2.1 Counterfactuals Generation

To address the interpretability issues exhibited in other metrics, we explore counterfactual explanations as an alternative. The common approach for modern counterfactual generation has been to structure it as an optimization problem, aiming to find the minimal changes to certain features of an input that will alter the final output while keeping most features constant. This method aims to compute human-understandable and realistic counterfactuals effectively, addressing the need for easy comprehension by human observers (Wachter et al., 2017).

However, due to the high dimensionality and discrete nature of text data, this approach has its limitations. First and foremost, formulating the task of minimizing the distance between two inputs as an optimization problem is challenging because it necessitates calculating gradients for a discrete input (Belinkov and Glass, 2019). Moreover, creating traditional counterfactuals that change the output class becomes increasingly difficult with longer text inputs. For instance, consider a lengthy movie review from a reviewer who passionately loves a particular movie. If we replace only a few words or tokens with high positive sentiments with negative sentiments, the overall sentiment classification of the review is unlikely to change, thus rendering "flipping the class" unrealistic in such cases.

To address these challenges, we propose *Soft Counterfactuals*, a new class of counterfactuals where a final output change is not necessarily required. In this relaxed definition, instead of finding "the closest possible world," we provide a diverse range of relevant and insightful "close possible worlds" (Wachter et al., 2017). Soft Counterfactuals allow for granular insights, such as detecting subtle changes in feature values that influence outcome probabilities, rather than requiring a complete shift from one outcome to another. Furthermore, without relying on an optimization function to minimize the distance, we utilize the power of the Transformer to ensure that these counterfactuals are realistic and contextually appropriate (Vaswani et al., 2017).

Another challenge in generating counterfactuals is resource constraints. Previous studies have explored human-generated counterfactuals for text data, but the high cost makes this approach impractical for large-scale operations. On average, workers spend 4 to 5 minutes revising a single input, and they are still prone to systematic omissions (Kaushik et al., 2019; Wu et al., 2021).

To overcome this obstacle, researchers have explored automated approaches, particularly leveraging the advancements in large language models (LLMs) that have gained traction in recent years. Notable examples include Wu et al. (2021), who introduced *Polyjuice*, a fine-tuned GPT-2 model trained on multiple datasets with paired sentences. Polyjuice employs control codes (e.g., negation, lexical changes) and a fill-in-the-blank structure to generate specific types of counterfactuals. Similarly, Robeer et al. (2021) created realistic counterfactuals using a Counterfactual GAN architecture. These methods and their counterparts often require auxiliary models and/or training data, which is not always feasible due to the lack of task-specific datasets or the resource-intensive nature of fine-tuning models (Bhattacharjee et al., 2024).

Bhattacharjee et al. (2024) proposed an alternative, utilizing zero-shot state-of-the-art LLMs to generate counterfactuals, but this approach can raise further concerns about the interpretability of the counterfactual generation process. Despite significant strides in generating realistic and high-quality counterfactuals with LLMs, most efforts have been directed towards using counterfactuals as a data augmentation tool. However, when the goal is to create standardized evaluation metrics for XAI techniques on text classification problems, the task becomes more straightforward.

In SCENE, we do not require counterfactuals to be perfect paraphrased representations of the original instances. Instead, we focus on token-based substitutions and compare the outputs holistically. Therefore, an off-the-shelf (zero-shot) *BERT model for masked language modeling* (*BertForMaskedLM*) is sufficient for the task (Devlin et al., 2018; Ribeiro et al., 2020). This approach is computationally efficient, requires no further fine-tuning, and can leverage GPU advantages.

In summary, SCENE is an approach that uses masked language modeling (MLM) for token-based substitution. We select and mask certain number of tokens, then replace them with likely alternatives to craft Soft Counterfactuals using a zero-shot BertForMaskedLM. This approach allows SCENE to measure *faithfulness*, as defined by Atanasova (2024), which refers to whether an explanation technique's outputs are faithful to the model's inner workings and not based on arbitrary choices. Additionally, SCENE measures *confidence indication*, as defined by Atanasova (2024), which refers to whether a token contributed significantly to the prediction.

### 3.2.2 Counterfactual Evaluation

The flexibility of Soft Counterfactuals offers convenience in generating realistic counterfactuals for text data but also poses challenges in evaluating the results. In this paper, we adopt the *Validity$_{soft}$* and *Counterfactual Evaluation Score$_{soft}$ (C$_{soft}$)* formulations defined by Ge et al. (2021) to evaluate XAI techniques' ability to provide relevant saliency explanations. In other words, these formulations assess the techniques' ability to identify tokens with high significance.

$$Validity_{soft} = \frac{1}{N}\sum_{n=1}^{N}\frac{1}{K}\sum_{k=1}^{K}\left(p(\widehat{y_n} \mid x_n) - p(\widehat{y_{n,k}} \mid x_{n,k}^{cf})\right)$$

Where:

> *N* is the number of original instances
> *K* is the number of Soft Counterfactuals generated for one instance
> $p(\widehat{y_n}|x_n)$ is the probability of the predicted label given the original input for instance ***n***
> $p\left(\widehat{y_{n,k}}\middle|x_{n,k}^{cf}\right)$ is the probability of the predicted label given the Soft Counterfactual input for instance ***n***

Validity$_{soft}$, inspired by the commonly used Validity metric for counterfactual evaluations, measures the average change in probabilities from the original predictions to the predictions made with counterfactual inputs for the same class. This metric assesses the extent of probability shifts—the higher the score, the better the method is at selecting the significant tokens (Mothilal et al., 2020; Ge et al., 2021).

$$C_{soft} = \frac{1}{N}\sum_{n=1}^{N}\frac{\sum_{k=1}^{K}\left(p(\widehat{y_n} \mid x_n) - p(\widehat{y_{n,k}} \mid x_{n,k}^{cf})\right)}{\sum_{k=1}^{K}\text{dist}(x_n, x_{n,k}^{cf})}$$

Where:

> *N* is the number of original instances
> *K* is the number of Soft Counterfactuals generated for one instance
> $p(\widehat{y_n}|x_n)$ is the probability of the predicted label given the original input for instance ***n***
> $p\left(\widehat{y_{n,k}}\middle|x_{n,k}^{cf}\right)$ is the probability of the predicted label given the Soft Counterfactual input for instance ***n***
> $\text{dist}\left(x_n, x_{n,k}^{cf}\right)$ is the distance between the original instance and its Soft Counterfactuals

C$_{soft}$ builds on the foundation of Validity$_{soft}$ by incorporating a distance function to evaluate the faithfulness of the saliency explanations (Ge et al., 2021). For the distance function, SCENE uses *Universal Sentence Encoder*, proposed by Cer et al. (2018). In this approach, the encoder (BERT-uncased for this paper) employs attention mechanisms to generate context-aware representations of words in a sentence, taking into account both the sequence and the identities of surrounding words. These context-aware word representations are then averaged to produce a sentence-level embedding (Cer et al., 2018). The higher the $C_{soft}$ score, the better the method is at providing faithful and informative explanations.

It is worth noting that $Validity_{soft}$ and $C_{soft}$ are designed for binary classification tasks and may not fully capture the behavior of counterfactual outputs in multi-class classification scenarios. For instance, an increase in $p\left(\widehat{y_{n,k}^{cf}} \mid x_{n,k}^{cf}\right)$ does not necessarily lead to a decrease in the original output $p(\widehat{y_{n,k}} \mid x_{n,k}^{cf})$; both probabilities can increase while the probabilities of other classes decrease (Ge et al., 2021).

## 4 Experimental Design

This section presents the experimental design of SCENE and the methods used to measure its performance. The experiment involves several stages: extraction of significant tokens, generation of Soft Counterfactuals, and computation of evaluation metrics.

### 4.1 Token Extraction

To identify the most significant tokens from the XAI results, SCENE extracts the top tokens based on their importance weights for each technique, respectively. Initially, the test set is passed through various XAI methods to ensure that attributions for all tokens are computed and saved for each method. Subsequently, the text is tokenized, with the first and last tokens removed to focus on the core content. The tokenized text is then paired with corresponding weights derived from three possible methods: mean, L2, or direct weighting. To ensure clarity, subword tokens indicated by '##' and their preceding tokens are filtered out. Additionally, non-alphabetic tokens are excluded unless they are single-character words such as 'I' or 'a'. The remaining tokens are sorted by their importance weights in descending order, and the top $V$ tokens are selected. This approach allows for a systematic ranking of the most relevant tokens, providing a framework for analyzing the saliency explanations generated by ten XAI methods across three different neural model architectures: CNN, RNN, and Transformer.

We selected ten XAI techniques to represent the most prominent model-agnostic XAI methods as of 2024. These include perturbation-based approaches such as LIME and SHAP, as well as backpropagation-based techniques like *Integrated Gradients* and *Guided Backpropagation*. It is worth noting that many XAI methods use gradients usually are not adapted for text

---

**Input**: $x_n$, $\mathcal{F}$, $\mathcal{E}_u$, $V$, $K$
**Output**: $x_{n,k}^{cf}$, $\mathcal{F}(x_{n,k}^{cf})$, $e_u^i$, $Validity_{soft}$, $C_{soft}$

1. $\widehat{y_n} \leftarrow p(\widehat{y_n} \mid x_n) \leftarrow \mathcal{F}(x_n)$
2. $e_u^i \leftarrow \mathcal{E}_u(x_n, \widehat{y_n})$
3. Identify top $V$ significant tokens in $x_n$ using $e_u^i$
4. **Foreach** $k$ in $1, \dots, K$ **do**
       Generate Soft Counterfactual $x_{n,k}^{cf}$ by replacing $V$ significant tokens using BertForMaskedLM
   **end**
5. $p(\widehat{y_{n,k}} \mid x_{n,k}^{cf}) \leftarrow \mathcal{F}(x_{n,k}^{cf})$
6. $dist(x_n, x_{n,k}^{cf}) \leftarrow USE(x_n) - USE(x_{n,k}^{cf})$
7. **Foreach** $n$ in $1, \dots, N$ **do**
       Calculate $\frac{1}{K} \sum_{k=1}^{K} \left( p(\widehat{y_n} \mid x_n) - p(\widehat{y_{n,k}} \mid x_{n,k}^{cf}) \right)$
       Calculate $\frac{\sum_{k=1}^{K} \left( p(\widehat{y_n} \mid x_n) - p(\widehat{y_{n,k}} \mid x_{n,k}^{cf}) \right)}{\sum_{k=1}^{K} dist(x_n, x_{n,k}^{cf})}$
   **end**
8. Return $Validity_{soft}$ and $C_{soft}$ for each $\mathcal{E}_u \leftarrow \frac{1}{N} \sum_{n=1}^{N} step\ 7$

Algorithm 1: Pseudocode for SCENE

explanations. For this reason, the attributions are calculated as a function of the embeddings and not the input tokens. The techniques include:

**LIME**: Proposed by Ribeiro et al. (2016), *Local Interpretable Model-agnostic Explanations (LIME)* creates locally faithful approximations around a specific prediction with a simpler model that is easier to interpret.

**SHAP**: Proposed by Lundberg and Lee (2017), *SHapley Additive exPlanations (SHAP)* applies Shapley values, concepts from cooperative game theory, to measure the contribution of individual features.

**Gradient X Input**: Proposed by Shrikumar et al. (2017), *Gradient X Input* serves as a straightforward method for visualizing each input feature's contribution to the output by multiplying the gradient of the output with respect to each input feature by the value of the input feature itself.

**Grad L2 Norm**: Closely related to Gradient X Input, *Gradient L2-norm* computes the L2-norm of the gradient of the model's output with respect to the inputs, which indicates the sensitivity of the model to changes in the input embeddings (Munn and Pitman, 2022).

**Integrated Gradients**: Proposed by Sundararajan et al. (2017), *Integrated Gradients* calculates the attribution to each input feature by multiplying the integrated gradient along that dimension by the difference between the actual input and the baseline.

***Saliency***: Proposed by Simonyan et al. (2013), *Saliency* creates a saliency map that highlights the regions of an image most influential to the classification decision of a neural network by utilizing a single backpropagation pass to calculate the gradients.

***Guided Backpropagation***: Proposed by Springenberg et al. (2014), *Guided Backpropagation* visualizes the parts of an image that most activate a given neuron in a neural network, combining the standard backpropagation approach with an additional guidance mechanism to provide clearer visualizations.

***Guided GradCAM***: Proposed by Selvaraju et al. (2017), *Guided Grad-CAM* is a variation of *Gradient-weighted Class Activation Mapping (Grad-CAM)* that generates coarse localization maps by computing the gradient of the target concept's score with respect to the feature maps of the last convolutional layer. This method highlights important regions in the image.

***DeepLIFT***: Proposed by Shrikumar et al. (2017), *Deep Learning Important FeaTures (DeepLIFT)* compares the activation of each neuron to its "reference activation" and assigns contribution scores based on this difference. This method backpropagates the contributions of all neurons in the network to every feature of the input to explain the output of neural networks.

***Deconvolution***: Proposed by Zeiler et al. (2014), *Deconvolution* maps the feature activations of a neural network back to the input pixel space by attaching a *deconvolutional network (deconvnet)* to each layer. This method allows for the visualization of the features learned by the neural network.

### 4.2 Creating Soft Counterfactuals

To generate Soft Counterfactuals with relevant and informative token replacements, SCENE masks the important tokens identified by XAI techniques and substitutes them with alternative words. These previously identified important tokens are masked, and the masked text is then re-tokenized. Using a zero-shot BertForMaskedLM model, we predict possible replacements for the masked tokens, filtering out non-alphabetic words and those identical to the original tokens. From the top predictions, we randomly select a subset to generate $K$ Soft Counterfactual samples. This process results in a collection of Soft Counterfactual texts, each with specified replacements for the masked tokens, enabling an in-depth analysis of the impact of token-level changes on model predictions. Additionally, a random seed is used to ensure reproducibility and a diverse range of Soft Counterfactuals.

SCENE generates $K$ Soft Counterfactuals where $V$ significant tokens are replaced by likely alternatives. In our experiments, the parameters were set to $K = 10$ and $V = 5$.

Table 1: A Simplified Example of SCENE Generation of Soft Counterfactuals When $V = 2$

| Original Sentence | I love this movie, the cast was great. |
|---|---|
| Identify Significant tokens | I [love] this movie, the cast was [great]. |
| Mask the Significant tokens | I [MASK] this movie, the cast was [MASK]. |
| Soft Counterfactual 1 | I [like] this movie, the cast was [awesome]. |
| Soft Counterfactual 2 | I [hate] this movie, the cast was [awful]. |
| Soft Counterfactual 3 | I [dislike] this movie, the cast was [lacking]. |
| Soft Counterfactual 4 | I [enjoy] this movie, the cast was [mediocre]. |
| … | … |

### 4.3 Evaluation of Soft Counterfactuals

Finally, SCENE computes the Validity$_{soft}$ and C$_{soft}$ (defined in 3.2.2) for the ten selected XAI methods across three different model architectures. We then present these results with two other metrics—infidelity and human agreements—to provide a comparison with SCENE.

Additionally, two possible ways to summarize these results are provided: the average of each token embedding attribution and the L2 norm. We also record the average time each XAI methodology spends on each test dataset instance, including both the inference and the calculation of attributions by the XAI methodology.[3]

## 5 Results

### 5.1 CNN (Table 2):

For the CNN architecture, the metrics reveal different strengths for each technique. In terms of Human Agreement, Saliency (both Mean and L2) performs well in capturing important attributes, making it more aligned with human interpretation of the model's decisions. When measuring Infidelity, Deconvolution (both Mean and L2) outperformed others by remaining more stable after perturbations, indicating a higher fidelity to the model's behavior.

---
[3] The times are measured in the following hardware setting: an L4 GPU with 53.0 GB system RAM, 22.5 GB GPU RAM, and 201.2 GB disk space.

* Best results highlighted. Time measured in seconds by instance of text. Up/down arrows denote higher/lower values implying better performance for each metric.

Table 2: Explainability Analysis for the CNN Architecture

| XAI Method | Human Agreement ↑ | Infidelity ↓ | Validity$_{soft}$ ↑ | C$_{soft}$ ↑ | Average Time ↓ |
|---|---|---|---|---|---|
| Lime - mean | 0.267737 | 0.000031 | 0.000011 | 0.014873 | 17.303985 |
| Lime - L2 | 0.270292 | 0.000031 | 0.000014 | 0.016460 | 17.303985 |
| SHAP - mean | 0.276594 | 0.000072 | 0.000456 | 0.435556 | 0.004722 |
| SHAP - L2 | 0.281154 | 0.000072 | 0.000105 | 0.109892 | 0.004722 |
| Input x Gradient - mean | 0.280178 | 0.000069 | 0.000475 | 0.485188 | 0.002712 |
| Input x Gradient - L2 | 0.283557 | 0.000069 | 0.000096 | 0.083646 | 0.002712 |
| Grad L2 Norm | 0.283840 | 0.000000 | -0.000027 | 0.061930 | 0.003254 |
| Integrated Gradients - mean | 0.280203 | 0.000065 | 0.000474 | 0.481461 | 0.071171 |
| Integrated Gradients - L2 | 0.283486 | 0.000065 | 0.000099 | 0.094937 | 0.071171 |
| Saliency - mean | 0.283703 | 0.000069 | 0.000061 | 0.064538 | 0.002973 |
| Saliency - L2 | 0.283840 | 0.000069 | 0.000063 | 0.061930 | 0.002973 |
| Guided Backpropagation - mean | 0.267827 | 0.000000 | -0.000027 | -0.039826 | 0.002672 |
| Guided Backpropagation - L2 | 0.283840 | 0.000000 | 0.000063 | 0.061930 | 0.002672 |
| Guided GradCAM - mean | 0.270946 | 0.000032 | 0.000056 | 0.064530 | 0.005678 |
| Guided GradCAM - L2 | 0.283157 | 0.000032 | 0.000065 | 0.067863 | 0.005678 |
| DeepLIFT - mean | 0.280178 | 0.000064 | 0.000475 | 0.485188 | 0.004503 |
| DeepLIFT - L2 | 0.283557 | 0.000064 | 0.000096 | 0.083646 | 0.004503 |
| Deconvolution - mean | 0.267827 | 0.000000 | -0.000027 | -0.039826 | 0.002651 |
| Deconvolution - L2 | 0.283840 | 0.000000 | 0.000063 | 0.061930 | 0.002651 |

Table 3: Explainability Analysis for the RNN Architecture

| XAI Method | Human Agreement ↑ | Infidelity ↓ | Validity$_{soft}$ ↑ | C$_{soft}$ ↑ | Average Time ↓ |
|---|---|---|---|---|---|
| Lime - mean | 0.259160 | 0.000281 | 0.000013 | 0.002648 | 79.395420 |
| Lime - L2 | 0.259905 | 0.000281 | 0.000016 | 0.007422 | 79.395420 |
| SHAP - mean | 0.290211 | 0.000321 | 0.000076 | 0.099034 | 0.067996 |
| SHAP - L2 | 0.377127 | 0.000321 | 0.000084 | 0.146383 | 0.067996 |
| Input x Gradient - mean | 0.294600 | 0.000282 | 0.000085 | 0.119695 | 0.015707 |
| Input x Gradient - L2 | 0.383988 | 0.000282 | 0.000068 | 0.123152 | 0.015707 |
| Grad L2 Norm | 0.384262 | 0.000000 | 0.000074 | 0.097952 | 0.036744 |
| Integrated Gradients - mean | 0.315418 | 0.000296 | 0.000095 | 0.143029 | 0.477190 |
| Integrated Gradients - L2 | 0.378088 | 0.000296 | 0.000066 | 0.124319 | 0.477190 |
| Saliency - mean | 0.377780 | 0.000576 | 0.000076 | 0.104001 | 0.015678 |
| Saliency - L2 | 0.384262 | 0.000576 | 0.000074 | 0.097952 | 0.015678 |
| Guided Backpropagation - mean | 0.306182 | 0.000000 | 0.000039 | 0.066461 | 0.015696 |
| Guided Backpropagation - L2 | 0.384262 | 0.000000 | 0.000074 | 0.097952 | 0.015696 |
| Guided GradCAM - mean | . | . | . | . | . |
| Guided GradCAM - L2 | . | . | . | . | . |
| DeepLIFT - mean | 0.294146 | 0.000284 | 0.000076 | 0.124642 | 0.031402 |
| DeepLIFT - L2 | 0.384053 | 0.000284 | 0.000068 | 0.123152 | 0.031402 |
| Deconvolution - mean | 0.306182 | 0.000000 | 0.000039 | 0.066461 | 0.015651 |
| Deconvolution - L2 | 0.384262 | 0.000000 | 0.000074 | 0.097952 | 0.015651 |

* Guided GradCAM requires an internal embedding layer, which is not present in the RNN structure used in this paper, hence it is omitted.

Table 4: Explainability Analysis for the Transformer Architecture

| XAI Method | Human Agreement ↑ | Infidelity ↓ | Validity$_{soft}$ ↑ | C$_{soft}$ ↑ | Average Time ↓ |
|---|---|---|---|---|---|
| Lime - mean | 0.266223 | 0.018785 | 0.031650 | 0.004013 | 116.532945 |
| Lime - L2 | 0.268752 | 0.018785 | -0.035652 | -0.013925 | 116.532945 |
| SHAP - mean | 0.283459 | 0.017712 | 0.204560 | 0.045396 | 4.603704 |
| SHAP - L2 | 0.286695 | 0.017712 | 0.328891 | 0.096335 | 4.603704 |
| Input x Gradient - mean | 0.306472 | 0.017102 | 0.188209 | 0.032034 | 0.972698 |
| Input x Gradient - L2 | 0.385673 | 0.017102 | 0.623118 | 0.144318 | 0.972698 |
| Grad L2 Norm | 0.395961 | 0.004041 | 0.685125 | 0.133388 | 6.338917 |
| Integrated Gradients - mean | 0.370959 | 0.016783 | 1.251692 | 0.416456 | 43.211839 |
| Integrated Gradients - L2 | 0.356946 | 0.016783 | 0.497796 | 0.094711 | 43.211839 |
| Saliency - mean | 0.395766 | 0.035320 | 0.644862 | 0.108248 | 0.886486 |
| Saliency - L2 | 0.395961 | 0.035320 | 0.685125 | 0.133388 | 0.886486 |
| Guided Backpropagation - mean | 0.304621 | 0.004266 | 0.234217 | 0.084060 | 0.980672 |
| Guided Backpropagation - L2 | 0.395961 | 0.004266 | 0.685125 | 0.133388 | 0.980672 |
| Guided GradCAM - mean | 0.309104 | 0.016006 | 0.547888 | 0.182488 | 1.611886 |
| Guided GradCAM - L2 | 0.388786 | 0.016006 | 0.623473 | 0.171880 | 1.611886 |
| DeepLIFT - mean | 0.307792 | 0.020099 | 0.288237 | 0.063594 | 1.708244 |
| DeepLIFT - L2 | 0.386062 | 0.020099 | 0.721534 | 0.239873 | 1.708244 |
| Deconvolution - mean | 0.304621 | 0.004205 | 0.234217 | 0.084060 | 1.060891 |
| Deconvolution - L2 | 0.395961 | 0.004205 | 0.685125 | 0.133388 | 1.060891 |

SHAP (both Mean and L2) showed strength in terms of the counterfactual metrics but exhibited high Infidelity and had a higher time cost, which may limit its practicality in real-time applications. Finally, gradient-based methods demonstrated greater efficiency regarding the time metric, providing faster explanations compared to other techniques.

### 5.2 RNN (Table 3):

Among the evaluated XAI techniques for RNN, Saliency (L2) stands out with the highest MAP at 0.3843, surpassing most other methods in this metric. Most L2 methods show similar MAP scores around 0.384, indicating comparable precision performance when considering L2 regularization. While most methods demonstrate average processing times under 0.05, LIME (both Mean and L2) requires significantly more time. Additionally, Saliency (both Mean and L2) exhibit slightly higher average infidelity (0.0006), suggesting they may be less stable in generating consistent explanations compared to other techniques.

In terms of Validity$_{soft}$, Integrated Gradients (mean) show the highest values, indicating their effectiveness in significantly altering the probability distributions when counterfactuals are introduced. For C$_{soft}$, SHAP (L2) performs well, suggesting it provide more faithful explanations.

### 5.3 Transformer (Table 4):

XAI methods applied to a BERT model reveal that Grad L2 Norm, Guided Backpropagation (L2), and Deconvolution (L2) offer the best performance, characterized by high Human Agreement, low Infidelity, strengths in Validity$_{soft}$ metrics, and reasonable average time. In contrast, methods such as LIME (both Mean and L2) performs poorly due to low MAP, high infidelity, low Validity$_{soft}$, low C$_{soft}$, and high average time.

### 5.4 SCENE's Performance

Table 5: Spearman's Correlation ($\rho$) when Comparing SCENE and Infidelity to Human Agreement

|  | Validity$_{soft}$ ↑ | C$_{soft}$ ↑ | Infidelity ↓ |
| --- | --- | --- | --- |
| Transformer | 0.865368 | 0.689991 | -0.203982 |
| RNN | 0.129596 | 0.164077 | -0.059564 |
| CNN | 0.140895 | 0.141083 | 0.088265 |

While human annotations are not definitive, for the purposes of this study, we consider them as the ground truth. We assess the performance of SCENE by comparing its Validity$_{soft}$ and C$_{soft}$ metrics, along with Infidelity, against the ground truth across three neural architectures using Spearman's correlation. Spearman's correlation is a statistical measure that evaluates the strength and direction of a monotonic relationship between two ranked variables.

Our initial empirical analysis reveals that SCENE metrics generally exhibit a stronger Spearman's correlation with the ground truth compared to Infidelity across all three architectures. Specifically, the correlation observed in the Transformer architecture is particularly strong, suggesting superior performance in this model. Conversely, the correlations in RNN and CNN architectures are weaker, indicating that SCENE's effectiveness may be architecture-dependent.

## 6 Conclusion

This paper presented SCENE (Soft Counterfactual Evaluation for Natural language Explainability), a novel approach that leverages LLMs to generate Soft Counterfactuals in a zero-shot manner. By focusing on token-based substitutions and employing a zero-shot BertForMaskedLM model, SCENE effectively circumvents the challenges associated with high-dimensional text data and the need for extensive fine-tuning. This streamlined approach not only ensures the generation of contextually appropriate and semantically meaningful Soft Counterfactuals but also maintains computational efficiency.

The introduction of the Validity$_{soft}$ and C$_{soft}$ metrics within SCENE provides a standardized framework for evaluating the effectiveness of XAI techniques in identifying significant tokens. These metrics, inspired by established concepts in causal inference, offer a nuanced understanding of the impact of token-level changes on model predictions, thus validating the faithfulness of the saliency explanations generated by different XAI methods.

The empirical results from applying SCENE across various model architectures, including CNN, RNN, and Transformer, indicate that different XAI techniques exhibit distinct strengths and limitations when applied to these models. For example, Saliency methods provide accurate insights while being computationally efficient, whereas Deconvolution ensures high fidelity to the model's behavior. Furthermore, SCENE demonstrates a strong Spearman's correlation with human agreement in the Transformer architecture when ranking the effectiveness of XAI techniques, and it consistently outperforms Infidelity across all tested architectures. However, SCENE's performance is less pronounced in RNN and CNN models, suggesting that its effectiveness may vary depending on the underlying model architecture. These findings are crucial for practitioners seeking to select the most appropriate XAI technique and model architecture, based on specific requirements such as interpretability, computational efficiency, and robustness to input perturbations.

In conclusion, SCENE represents an alternative in the evaluation of XAI techniques for NLP tasks. By addressing the limitations of existing methods and leveraging the capabilities of LLMs, SCENE provides a powerful tool for generating realistic counterfactuals and assessing the effectiveness of XAI methods.

# 7 Future Work

Future work will focus on advancing SCENE by refining evaluation metrics to better capture the behavior of counterfactual outputs, especially in multi-class settings, addressing the unique challenges posed by complex label structures and interactions. Additionally, integrating SCENE with diverse NLP tasks such as named entity recognition, machine translation, and question answering will broaden its applicability. Enhanced counterfactual generation techniques, including exploring state-of-the-art LLMs and fine-tuning for specific domains, may further improve the quality and relevance of generated counterfactuals. By addressing these areas, we aim to enhance the robustness, flexibility, and impact of SCENE, contributing to the development of more interpretable, transparent, and accountable AI systems.

# 8 Acknowledgments

We would like to express our gratitude to Pamela González Muñiz and Joselo Peña Contreras for their assistance in facilitating the experimentation process and testing SCENE on the CNN and Transformer structures. We also extend our thanks to Dr. Batuhan Gundogdu and Dr. Yuri Balasanov for their helpful discussions and feedback.